\documentclass[lettersize,journal]{IEEEtran}
\usepackage{amsmath,amsfonts}
\usepackage{algorithmic}
\usepackage{array}
\usepackage[caption=false,font=normalsize,labelfont=sf,textfont=sf]{subfig}
\usepackage{textcomp}
\usepackage{stfloats}
\usepackage{url}
\usepackage{verbatim}
\usepackage{xcolor}
\usepackage{graphicx}

\usepackage{breakurl}
\usepackage[colorlinks=true,
            urlcolor=blue]{hyperref}
\usepackage[numbers]{natbib}
\usepackage{algorithm}
\usepackage{soul}
\usepackage{multirow}
\usepackage{orcidlink}
\usepackage{booktabs} 
\usepackage{lineno}
\usepackage{romannum}
\newcommand{\RN}[1]{%
  \textup{\uppercase\expandafter{\romannumeral#1}}%
}

\begin{document}

\title{Planning Safety Trajectories with Dual-Phase, Physics-Informed, and Transportation Knowledge-Driven Large Language Models}

\author{Rui GAN\hspace{-1.5mm}$^{~\orcidlink{0009-0007-1344-9703}}$,
      Pei LI*\hspace{-1.5mm}$^{~\orcidlink{0000-0002-7512-3705}}$,
      Keke LONG\hspace{-1.5mm}$^{~\orcidlink{0000-0002-9244-739X}}$,
      Junyi MA,
      Bocheng AN\hspace{-1.5mm}$^{~\orcidlink{0000-0003-2632-2748‬}}$,
      Junwei YOU,
      Keshu WU\hspace{-1.5mm}$^{~\orcidlink{0000-0002-8085-0990}}$ and
      Bin RAN\hspace{-1.5mm}$^{~\orcidlink{0000-0002-0845-8716}}$
\thanks{* Corresponding Author.}
\thanks{R. Gan, P. Li, K. Long, J. Ma, J. You, and B. Ran are all associated with the Department of Civil and Environmental Engineering at the University of Wisconsin-Madison, Madison, WI, 53706, USA (e-mail: rgan6@wisc.edu, pei.li@wisc.edu, klong23@wisc.edu, jma333@wisc.edu, jyou38@wisc.edu, and bran@wisc.edu).}

\thanks{K. Wu is affiliated with the Center for Geospatial Sciences, Applications, and Technology, as well as the Department of Landscape Architecture and Urban Planning, and the Zachry Department of Civil and Environmental Engineering, Texas A\&M University, College Station, TX, 77840, USA (email: keshuw@tamu.edu).}

\thanks{B. An is affiliated with the School of Transportation, Southeast University, Nanjing, 211189, China (email: anbc17@seu.edu.cn).}
}

\markboth{Journal of \LaTeX\ Class Files,~Vol.~14, No.~8, August~2021}%
{Shell \MakeLowercase{\textit{et al.}}: A Sample Article Using IEEEtran.cls for IEEE Journals}

\maketitle

\begin{abstract}
Foundation models have demonstrated strong reasoning and generalization capabilities in driving-related tasks, including scene understanding, planning, and control. However, they still face challenges in hallucinations, uncertainty, and long inference latency. While existing foundation models have general knowledge of avoiding collisions, they often lack transportation-specific safety knowledge. To overcome these limitations, we introduce LetsPi, a physics-informed, dual-phase, knowledge-driven framework for safe, human-like trajectory planning. To prevent hallucinations and minimize uncertainty, this hybrid framework integrates Large Language Model (LLM) reasoning with physics-informed social force dynamics. LetsPi leverages the LLM to analyze driving scenes and historical information, providing appropriate parameters and target destinations (goals) for the social force model, which then generates the future trajectory. Moreover, the dual-phase architecture balances reasoning and computational efficiency through its Memory Collection phase and Fast Inference phase. The Memory Collection phase leverages the physics-informed LLM to process and refine planning results through reasoning, reflection, and memory modules, storing safe, high-quality driving experiences in a memory bank. Surrogate safety measures and physics-informed prompt techniques are introduced to enhance the LLM's knowledge of transportation safety and physical force, respectively. The Fast Inference phase extracts similar driving experiences as few-shot examples for new scenarios, while simplifying input-output requirements to enable rapid trajectory planning without compromising safety. Extensive experiments using the HighD dataset demonstrate that LetsPi outperforms baseline models across five safety metrics. Ablation studies further confirm that the dual-phase design achieves superior results compared to direct few-shot approaches while significantly reducing inference time. Project page:\url{https://github.com/mcgrche/LetsPi--Planning-Safety-Trajectories-with-Dual-Phase-Physics-Informed-LLM}.
\end{abstract}


\section{Introduction}
\IEEEPARstart{S}{afety} is a fundamental concern in vehicle trajectory planning, which is crucial in various Intelligent Transportation Systems (ITS) applications. However, trajectory planning is a complicated task that requires understanding dynamic traffic environments, predicting the behavior of surrounding vehicles, and making real-time decisions under uncertainty. Recent advancements in Generative Artificial Intelligence (GAI), particularly Large Language Models (LLMs), offer new opportunities in this domain. Pre-trained on internet-scale datasets with hundreds of billions of parameters, LLMs can reason over complex scenarios, generalize across diverse situations, and produce context-aware decisions, making them a promising tool in trajectory planning.

Trajectory planning predicts a vehicle’s transition from one feasible state to the next while considering various factors, including vehicle dynamics, occupant comfort, lane boundaries, traffic rules, etc~\cite{katrakazas2015real}. Among the widely used approaches, model-based methods such as Model Predictive Control (MPC), rapidly-exploring random trees, and Bézier curves explicitly represent vehicle kinematics, control laws, and physical constraints~\cite{dixit2018trajectory,claussmann2019review}. However, model-based approaches are often highly specialized and struggle to balance multiple objectives or handle complex constraints, which limits their generalization to diverse real-world scenarios. To address these limitations, learning-based approaches have been developed, such as reinforcement learning, imitation learning, etc. These methods learn driving policies directly from data, enabling adaptability to complex environments with multiple objectives. Nonetheless, learning-based methods rely on large-scale and high-quality data, which can be difficult to obtain. Moreover, generalization remains a concern when applying these models to unseen situations. In addition, learning-based methods often lack interpretability, making it hard to understand or trust the learned policies in safety-critical applications like vehicle trajectory planning.

LLMs offer a promising solution to address these limitations~\cite{brown2020language}. With their strong reasoning capabilities and language-based framework, LLMs can naturally provide interpretable solutions and generalize to diverse scenarios. As a result, recent studies have started exploring their application in ITS, including autonomous driving~\cite{wen2023dilu,shao_lmdrive_2023,wang2023drivemlm,hu2024agentscodriver,mao_gpt-driver_2023}, signal optimization~\cite{movahedi_crossroads_2025}, safety analysis~\cite{wang2023accidentgpt,li2024driving}, and traffic object detection~\cite{zhu_vision-language_2025,hasan_vision-language_2024}. 


However, several challenges remain in planning safety trajectories with LLMs. First, LLMs possess general knowledge, including safety principles such as avoiding collisions. However, they often struggle to ensure that their decisions strictly satisfy complex safety constraints~\cite{wen2023dilu,wang2023drivemlm,mao_gpt-driver_2023,sharan_llm-assist_2023}. 
For example, although LLMs understand that vehicles should not collide, they may fail to recognize if a trajectory leads to unsafe situations such as conflicts and near-misses. Second, LLMs are prone to hallucination and can generate inconsistent outputs. This limitation stems from the inherent uncertainty in LLMs' outputs and the lack of reflection on whether their actions meet physical laws or safety requirements. Lastly, LLM-based models often require lengthy inference time to reason and generate outputs~\cite{hu2024agentscodriver,chen2024driving,sima2024drivelm}. This may prevent implementing these models in practical, real-time scenarios.

\begin{figure*}[bhtp]
  \begin{center}
  \includegraphics[width=0.8\textwidth]{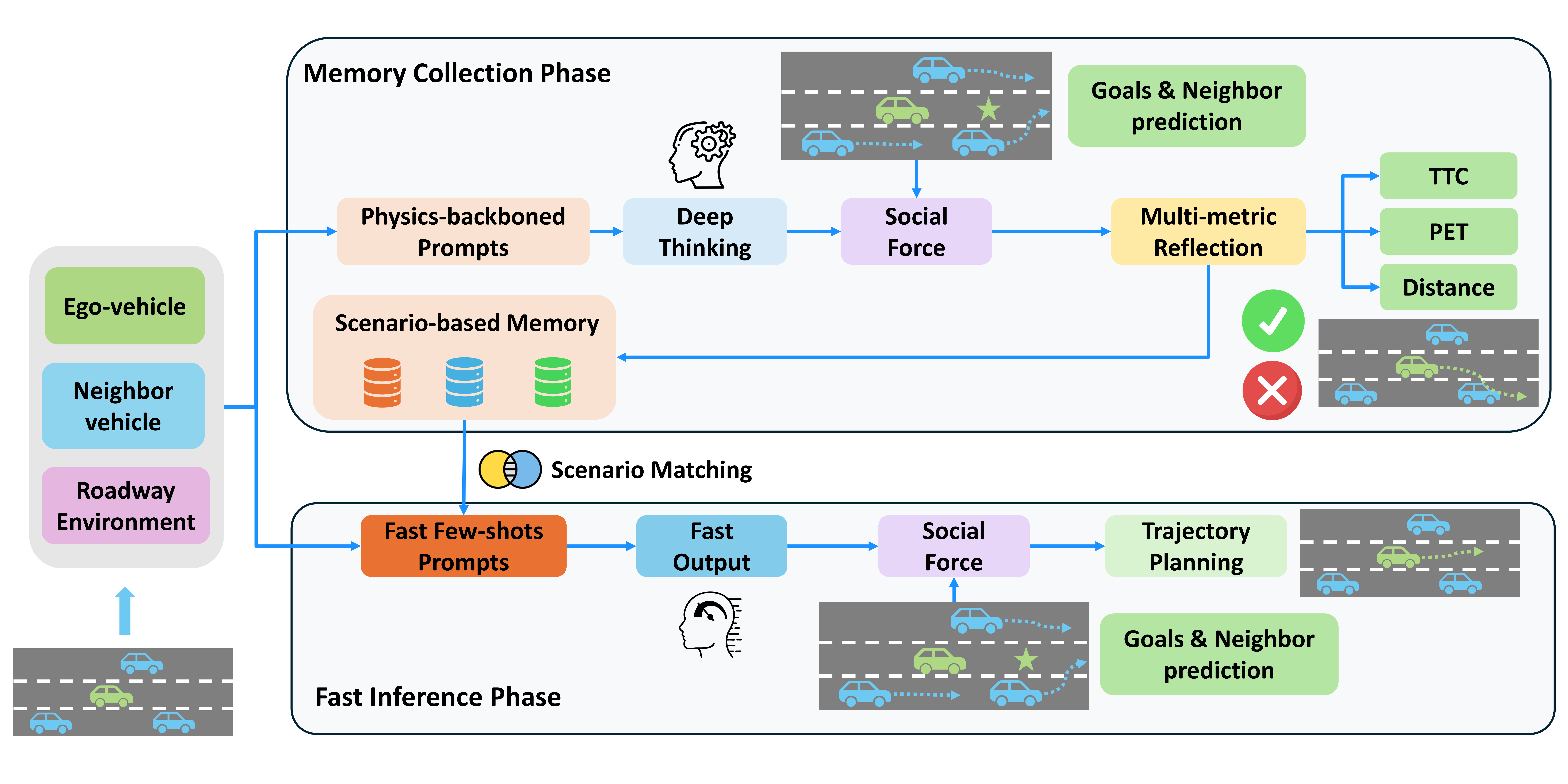}\\
  \caption{LetsPi, a dual-phase, physics-informed LLM architecture for safe trajectory planning. The memory collection phase builds the knowledge database through in-depth reasoning and reflection using multiple safety metrics. The fast inference phase distills knowledge from the memory database, enabling timely trajectory planning.}\label{fig1_framework}
  \end{center}
\end{figure*}

To address these challenges, we propose LetsPi, a novel dual-phase, physics-informed, and knowledge-driven LLM framework for safe trajectory planning (Fig.~\ref{fig1_framework}). The contributions of this paper are as follows:
\begin{itemize}
  \item We integrate the reasoning capability of LLM with the reliable physics dynamics of the social force model. By embedding explicit physical constraints into LLM reasoning, our framework facilitates accurate interpretation and scenario understanding, producing recommended social force parameters and adjusted goals. These recommendations are subsequently transformed into safe, realistic driving trajectories by the social force model.
  \item We design a dual-phase framework that balances in-depth reasoning with real-time performance. The memory collection phase builds a knowledge base through comprehensive reasoning, physics model incorporation, and multi-metric reflection. The fast inference phase distills knowledge from the memory through scenario matching, enabling timely trajectory planning. Experimental results indicate that the fast inference phase delivers inference speeds up to two times faster than the other phase, all while maintaining comparable safety performance. 
  \item Two prompt techniques are proposed. The structured, physics-informed prompt incorporates a detailed scenario description and physical driving knowledge. The fast prompt reduces explicit reasoning with few-shot examples, focusing on essential parameter retrieval.
  \item We introduce transportation safety knowledge into the reflection module using surrogate safety measures, including time-to-collision (TTC), post-encroachment time (PET), and minimum distance to collision. This design helps the LLM proactively assess and refine its outputs based on transportation safety knowledge.
\end{itemize}


The rest of the paper is organized as follows: Section \RN{2} reviews the roles and usage of LLMs in existing transportation research. Section \RN{3} details the architecture and essential components of the proposed framework. Section \RN{4} provides evaluation results of the framework. Section \RN{5} presents conclusions and potential improvements.

\section{Related Work}

Table~\ref{tab:summary} summarizes recent research on utilizing LLMs across various ITS applications. This section reviews existing studies from two aspects: the role assigned to LLMs and how LLMs are employed.

\subsubsection{Roles of LLMs} LLMs function either as standalone agents or as integral components of hybrid frameworks. In the first scenario, LLMs act as intelligent agents that process multiple data types, such as text, images, and instructions, to control autonomous vehicles (AVs)~\cite{wen2023dilu,wang2023drivemlm,hu2024agentscodriver},  plan trajectories~\cite{mao_gpt-driver_2023}, optimize traffic signals~\cite{movahedi_crossroads_2025}, or generate safety suggestions~\cite{wang2023accidentgpt,li2024driving}. Because standalone LLMs rely on general knowledge, they require extra components to refine their decisions for domain-specific tasks. These often include memory and reflection modules, which store past decisions and improve future outputs~\cite{wen2023dilu,hu2024agentscodriver,movahedi_crossroads_2025}. For example,~\citet{wen2023dilu} proposed a framework with three components, including reasoning, memory, and reflection. The framework includes a reasoning module that generates the driving policy by comprehending the environment using GPT 3.5. Second, a memory module stores the decision history or historical references to provide the reasoning module with experiences. Third, a reflection module is used to provide feedback on the driving policy and update the memory module.~\citet{hu2024agentscodriver} adopted a similar design while enhancing the reflection module with a reinforcement component. This new component contains an evaluator that assesses the decisions of LLMs and a reflector that helps the LLMs to learn from their historical mistakes. Similarly,~\citet{movahedi_crossroads_2025} designed an LLM-based actor-critic framework for adaptive traffic signal control. The actor agent analyzes real-time traffic conditions and generates signal phase plans, while the critic agent evaluates these decisions and updates a knowledge repository to refine future actions. Similar designs are applied in other applications. \citet{wang_accidentgpt_2023} have incorporated an update module that evaluates and refines LLM decisions for safety evaluation and prediction.

Instead of relying solely on LLMs, recent studies have integrated LLMs with other models and proposed hybrid frameworks~\cite{shao_lmdrive_2023,zhu_vision-language_2025,hasan_vision-language_2024,sharan_llm-assist_2023,long_vlm-mpc_2024,wang_empowering_2024,chen2024asynchronous}. For example,~\citet{sharan_llm-assist_2023} proposed a hybrid framework that combines an LLM with existing planners. This framework uses PDM-Closed and an intelligent driver model (IDM) as the base planner. The LLM is triggered to generate parameters for the base planner when its decisions are unsafe. Similarly,~\citet{long_vlm-mpc_2024} introduced a VLM-MPC framework that combines a vision-language model (VLM) with an MPC. The VLM generates parameters for the MPC, which executes control of the ego-vehicle.~\citet{wang_empowering_2024} leverages an LLM to provide decision support for an MPC-based controller. The LLM generates a target lane for the lower-level MPC controller with the reasoning. Moreover, LLMs were integrated with other models to enhance object detection and tracking in transportation.~\citet{zhu_vision-language_2025} proposed a vision-language object-tracking framework. This framework integrates CLIP-based feature extraction with interactive prompt learning, eliminating heavy modality fusion. Using three types of prompts, it enhances vision-language fusion and improves tracking efficiency and accuracy. Similarly,~\citet{hasan_vision-language_2024} leveraged vision-language models to detect distracted driving from naturalistic driving data. This study utilized pre-trained CLIPs to extract visual and text embeddings, which serve as inputs for a simple classification model. This framework enables fine-tuning the classifier without modifying CLIP itself, improving computational efficiency while maintaining strong performance in distracted driver detection.

\subsubsection{Usage of LLMs}

Prompt engineering and fine-tuning are the two most common methods for instructing LLMs. Prompt engineering communicates with LLMs to steer their behavior for desired outcomes without updating their weights~\cite{weng2023prompt}. Because of this advantage, most studies employed prompt engineering to guide LLMs in transportation applications as shown in Table~\ref{tab:summary}, using techniques including zero-shot learning (ZSL), few-shot learning (FSL), and chain-of-thought prompting (CoT). 

Zero-shot learning is the most basic prompt engineering technique and works by feeding the task text to the LLM and asking for results. The premise of zero-shot learning is that the LLM can generalize to different domain-specific tasks because of the diverse and extensive data they have been trained on. For instance,~\citet{li2024driving} indicated that LLMs can interpret traffic rules in the local driver handbook without additional training. The authors used GPT-4 with zero-shot learning to provide driving suggestions to drivers who are unfamiliar with local traffic rules. Differently, few-shot learning enhances model performance by providing a small set of high-quality input-output examples to guide the LLM toward the desired response pattern. The structure of these examples varies across studies, such as mapping driving descriptions to driving actions~\cite{wen2023dilu}, environment descriptions to control parameters~\cite{long_vlm-mpc_2024}, or driver instructions to vehicle actions~\cite{wang2023drivemlm}. However, the key assumption remains that LLMs learn better when given well-selected examples. This makes the selection of diverse and representative examples crucial for achieving optimal results~\cite{su2022selective}.

CoT is another effective technique in prompt engineering. It generates a sequence of short sentences to describe reasoning logics step by step and eventually leads to the final answer~\cite{wei2022chain}. This technique is commonly combined with ZSL or FSL to enhance LLM decision-making. For example,~\citet{wen2023dilu} designed a prompt that instructs the LLM by defining its role, output formats, and things to consider while making driving decisions. CoT is used to ask the LLM to generate a sequence of sentences that describe the step-by-step reasoning logic while making driving decisions. Based on the behavior and position of neighboring vehicles, the LLM first decides if the ego-vehicle can accelerate, then idle, and finally decelerate.~\citet{wang2023accidentgpt} utilizes CoT to evaluate safety in driving scenarios. The prompt includes images and text descriptions of the driving environment, and the LLM generates safety evaluations based on the input.~\citet{movahedi_crossroads_2025} used CoT to optimize traffic signal timing. The prompt contains step-by-step instructions for optimizing signals. The LLM first determines the number of roads that get the green light, followed by deciding which road or roads will get the green light. It then decides on right-turn permissions and the duration of each signal phase.

Fine-tuning (FT) aims to train the LLM on a specific dataset to adapt its parameters to the target task. This process requires additional computational resources and data, but may lead to significant improvements in performance. However, since training the entire LLM requires extensive computing resources, existing studies have adopted methods such as LoRA~\cite{hu2022lora} to insert a smaller number of new weights into the LLM and only these are trained. For example,~\citet{chen2024asynchronous} used LoRA to train the LLM using datasets that contain questions and answers for autonomous driving. Questions include descriptions about driving environments and the behavior of other vehicles, while answers include the ego-vehicle's behavior, such as acceleration, deceleration, etc. Fine-tuning is useful when the pre-trained LLMs lack domain-specific knowledge or perform poorly for specialized tasks. However, extensive, high-quality training data and computing resources are also required.

In summary, LLMs demonstrate strong reasoning and decision-making capabilities in various ITS applications due to the diverse and extensive knowledge they have acquired. Existing studies have suggested a multi-module design to employ LLMs. In particular, memory and reflection modules can significantly improve LLMs' performance by providing refined, high-quality few-shot examples. Moreover, hybrid frameworks that combine LLMs with other models are a promising research direction for expanding the applications of LLMs. However, several gaps remain in existing studies. The first is the lack of incorporation of physics-based knowledge into LLMs' reasoning process, which may lead to uncertainty and hallucination in its outputs. Second, the computing time is a major concern in implementing LLM-based solutions in real-world, real-time scenarios. Moreover, knowledge of driving safety is not comprehensively introduced into LLMs. The rate of collisions is commonly used, but it lacks the consideration of other critical events, including conflicts and near-misses. 


\begin{table*}[!t]
\caption{Summary of Studies using LLMs for ITS Tasks}
\centering
\begin{tabular}{llllll}
\hline
\textbf{Type} & \textbf{Study} & \textbf{Input} & \textbf{Output} & \textbf{LLM Instruction} & \textbf{Structure}\\
\hline
\multirow{9}{*}{Autonomous Driving} 
& \cite{wen2023dilu} & Text & Action & FSL, CoT & LLM\\
& \cite{mao_gpt-driver_2023} & Text & Trajectory & FSL, CoT, FT & LLM\\
& \cite{long_vlm-mpc_2024} & Image & Parameters & FSL, CoT & Hybrid\\
& \cite{sharan_llm-assist_2023} & Text & Parameters & FSL, CoT & Hybrid \\
& \cite{shao_lmdrive_2023} & Text & Action & ZSL & Hybrid\\
& \cite{wang2023drivemlm} & Vector, Text & Action & ZSL, CoT & LLM\\
& \cite{wang_empowering_2024} & Text & Parameters & CoT & Hybrid\\
& \cite{hu2024agentscodriver} & Text & Action & FSL, CoT & LLM\\
& \cite{chen2024asynchronous} & Text & Embeddings & FT & Hybrid\\
\hline
\multirow{3}{*}{Traffic Analysis} 
& \cite{li2024driving} & Image, Text & Driving suggestions & ZSL & LLM \\
& \cite{wang2023accidentgpt} & Image, Text & Safety evaluation & CoT & LLM\\
& \cite{movahedi_crossroads_2025} & Text & Signal timing & ZSL, CoT & LLM\\
\hline
\multirow{2}{*}{Object Detection}  
& \cite{zhu_vision-language_2025} & Image, Text & Embeddings & ZSL & Hybrid\\
& \cite{hasan_vision-language_2024} & Image, Text & Embeddings & ZSL & Hybrid\\
\hline
\end{tabular}
\label{tab:summary}
\end{table*}

\section{Methodology}
\subsection{Overviews}
Fig.~\ref{fig1_framework} presents the physics-informed, dual-phase and knowledge-driven LetsPi architecture for safe trajectory planning. This framework balances deep, context-rich reasoning with efficient, real-time decision-making with two phases: (1) \textbf{Memory Collection} and (2) \textbf{Fast Inference}. Specifically, the Memory Collection phase builds a transportation knowledge-based memory database, recording driving scenarios with validated parameters and trajectories. Using the physics-informed prompts, the LLM Engine returns social force model parameters and goal predictions by analyzing ego-vehicle intentions, neighbor interactions, and lane-specific environmental conditions. Outputs from the LLM are used by the social force model to plan ego-vehicle trajectories. Moreover, a reflection module is designed to assess these trajectories. This module evaluates trajectories with multiple safety metrics, including TTC, PET, and minimum distance to collision. Safe trajectories are stored in the memory database, while unsafe trajectories are refined by re-engaging the LLM Engine with additional safety prompts. On the other hand, the Fast Inference phase distills knowledge from the memory database through scenario matching. This module uses concise prompts to rapidly generate essential parameters, enabling timely trajectory planning.

\subsection{Physics-informed Prompt Generation}
\begin{figure*}[bhtp]
  \begin{center}
  \includegraphics[width=0.9\textwidth]{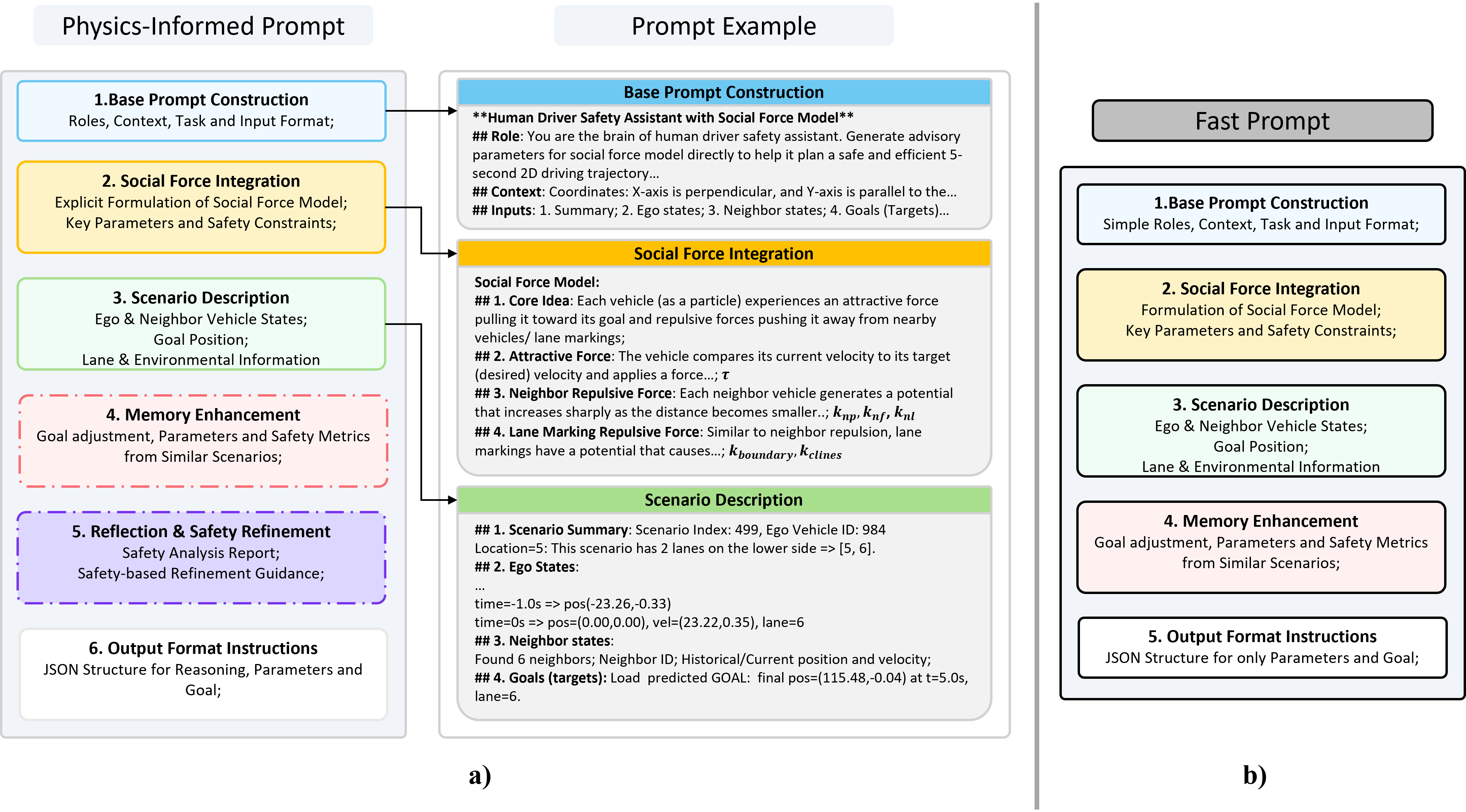}\\
  \caption{a) Physics-informed Prompt Architecture. Essential components include base prompt construction, social force integration, scenario description, and output instruction. Two optional prompts are used for memory enhancement and refinement. b) Fast Prompt Architecture. Without reflection and safety refinement,  Fast Prompt comprises simplified base prompt construction, social force integration, scenario description components and minimalistic output requirement with dedicated examples from memory.}\label{fig2_prompt}
  \end{center}
\end{figure*}

Ensuring safe and reliable trajectory planning requires a comprehensive understanding of vehicle dynamics and interactions between vehicles and roadway environments. We introduce a Physics-informed Prompt architecture depicted in Fig.~\ref{fig2_prompt} a) to achieve this. 

The Physics-informed Prompt has four essential and two optional components. First, the base prompt provides a role definition and context for the trajectory planning. These trivial definitions are effective in guiding the LLM engine for domain-specific tasks~\cite{wen2023dilu}. 
Second, the social force integration prompt is introduced to integrate the physics model with the LLM engine. This design explicitly models critical interactions defined in the social force model: attractive forces directed by $\tau$ represent driver goals, while repulsive forces among neighboring vehicles are governed by distinct interaction parameters $k_{np}$, $k_{nf}$, $k_{nl}$. Environmental influences, such as lane markings, are accounted for via boundary constraints defined by $k_{boundary}$, $ k_{cline}$. This explicit representation grounds the model’s parameter selection firmly in real-world physical principles, significantly reducing the abstraction and ambiguity typically inherent in pure learning-based approaches. Third, prompts are further enriched with comprehensive, scenario-specific contextual inputs, including historical and real-time vehicle states, inferred driver intentions, and detailed lane-marking configurations. Fourth, output format instructions are introduced to ensure that the outputs from the LLM conform to a standardized JSON format. The outputs contain selected parameters, explanatory rationales from the LLM, and predicted trajectories. Lastly, two optional components, including memory enhancement and reflection, are utilized for constructing the memory base and refining the LLM decision. More details about them will be provided in \textbf{Section \MakeUppercase{\romannumeral 5}-E}.

In summary, by explicitly incorporating both the social force model interpretation and a detailed description of the driving scenario into structured, physics-informed prompts, with clearly articulated tasks and strict output constraints, the LLM reliably generates essential social force model parameters. Moreover, the LLM provides explicit reasoning and detailed explanations justifying its selection, transforming the LLM into an interpretable, physics-based parameter generator tailored specifically for driving scenarios.



\subsection{LLM Reasoning and Inference}
Using the physics-informed prompt, the reasoning process of the LLM can be represented as:

\begin{equation}
\theta_{t=0} = f_{LLM}(E_{t=-h:t=0}, N_{t=-h:t=0}, G_f, L, P)
\end{equation}

The parameter vector at the current time step (t = 0) consists of six critical elements: $\tau$ (relaxation time controlling the convergence toward a desired velocity), $k_np$ (repulsion strength from preceding vehicles), $k_nf$ (repulsion strength from following vehicles), $k_nl$ (repulsion strength from lateral vehicles), $k_boundary$ (repulsion strength from lane boundaries), and $k_cline$ (attraction strength toward lane centerlines). These parameters collectively determine vehicle behavior within the social force model. The input $E_{-h:0}$ encompasses ego vehicle states from $h$ steps in the past to the current time 0, including historical trajectory points $(x_i, y_i, lane_i)$ for each time step $i$ from $-h$ to $-1$, and the current state $(x_0, y_0, v_{x,0}, v_{y,0}, lane_0)$ with position, velocity, and lane information. Similarly, $N_{-h:0}$ captures neighboring vehicle states over the same time horizon, with historical data $(x_{j,i}, y_{j,i}, lane_{j,i})$ for each vehicle $j$ at past time steps $i$, and current states $(x_{j,0}, y_{j,0}, v_{x,j,0}, v_{y,j,0}, lane_{j,0})$ for all $m$ neighbors. The goal state $G_f = (x_f, y_f)$ specifies the target destination at future horizon $f$, while $L$ provides lane marking information, including boundaries and center lines. Finally, $P$ represents the physics background knowledge from the social force model, serving as the physical backbone that grounds the LLM's reasoning in real-world vehicle dynamics.

Through analyzing these inputs, the LLM assesses scenario-critical details such as relative positioning and velocities among vehicles, lane configurations, potential lane-change maneuvers, current driving actions, and relevant safety considerations. It then applies physics-informed reasoning to interpret and calculate the social force model equations, including the attractive force guiding the vehicle toward the target velocity, proportional to the velocity difference and relaxation parameter $\tau$ (\textbf{Eq. (4)}); repulsive forces from neighboring vehicles, influenced by parameters $k_np, k_nf$, and $k_nl$, computed based on relative positions (\textbf{Eq. (6)}); and repulsive forces from lane markings that include boundary repulsion and centerline repulsion, modulated by parameters $k_boundary$ and $k_cline$. (\textbf{Eq. (7)}). 

Ultimately, leveraging this detailed and physically grounded reasoning process, the LLM outputs optimal values for the six parameters ($\tau, k_{np}, k_{nf}, k_{nl}, k_{boundary}, k_{cline}$). Each parameter is justified by explicit reasoning from the LLM, ensuring transparency, interpretability, and reliability in the generated trajectories.

\subsection{Physical Backbone - Social Force Model}
The social force model is based on a physics-inspired approach for modeling object trajectories, interpreting traffic participants as particles influenced by attractive and repulsive forces. Initially introduced to describe pedestrian movements and subsequently adapted for vehicular traffic scenarios~\cite{helbing1995social}, the social force model captures vehicle interactions by balancing a synthesis of goal-directed attractive forces and repulsive forces. Individual interaction is expressed mathematically through parametric force equations, with adjustable parameters precisely controlling the strength, sensitivity, and spatial range of each influence. This parameterized approach enables nuanced, adaptive modeling of complex vehicle dynamics across varied traffic contexts, ensuring both accuracy and physical plausibility of the generated trajectories. Our implementation of the social force module closely follows the formulation described in \cite{gan2024goal, yue2022human}, please refer to these sources for more comprehensive information.

The social force model employs Newtonian mechanics to describe the motion of vehicles in traffic contexts. Vehicles are treated as particles that follow Newton's second law, expressed through two fundamental equations. The acceleration at each time step is determined by summing all relevant forces (assuming a simplified unit mass):

\begin{equation}
\ddot{p}(t) = F_{\text{goal}}(t, p^T, p(t)) + F_{\text{rep}}(t, p(t), \Omega(t))
\end{equation}

In this formulation, the goal-directed attraction force $F_{\text{goal}}$ guides the vehicle toward its designated destination, and the repulsive force $F_{\text{rep}}$ ensures adequate separation from other vehicles and road boundaries. Following the acceleration calculation, the position and velocity for the next timestep are simultaneously updated via:

\begin{equation}
\begin{pmatrix} p(t + \Delta t) \\ \dot{p}(t + \Delta t) \end{pmatrix} =
\begin{pmatrix} p(t) \\ \dot{p}(t) \end{pmatrix} +
\Delta t \begin{pmatrix} \dot{p}(t) \\ \ddot{p}(t) \end{pmatrix}
\end{equation}

Through iterative computation, this model progressively generates comprehensive trajectories, where realistic vehicle behaviors naturally emerge from the continuous interplay between attractive destination-driven forces and repulsive safety-preserving forces within complex traffic interactions. In this iterative trajectory generation process, the social force model simultaneously integrates predicted future goals of the ego vehicle and predicted trajectories of surrounding vehicles, ensuring physically coherent and realistic vehicle movements. Specifically, we incorporate a transformer-based multi-goal prediction approach introduced in our previous work \cite{gan2024goal} to estimate potential ego-vehicle goals, whereas neighboring vehicles' trajectories are approximated using IDM for computational simplicity. Next, we provide a detailed introduction to the two core force components: the attraction force and the repulsion force.

\vspace{1em}
\noindent \textbf{Goal Attraction:} 
The goal attraction force models a vehicle's inherent driving intention, guiding it toward its target destination. At time \( t \), the vehicle's intended driving direction \( e^t \) is computed using the goal position \( p^T \) and current position \( p^t \) as \( e^t = \frac{p^T - p^t}{\|p^T - p^t\|} \). Unlike traditional static approaches, we dynamically adjust the desired speed \( v_0^t \) based on the vehicle’s proximity to its goal at every timestep: \( v_0^t = \frac{\|p^T - p^t\|}{(T-t)\Delta t} \). The desired velocity is thus given by \( v_{\text{des}}^t = v_0^t e^t \). Accordingly, the attraction force, facilitating smooth velocity adjustments toward the goal within relaxation time \( \tau \), is expressed as:

\begin{equation}
F_{\text{goal}} = \frac{1}{\tau}(v_{\text{des}}^t - \dot{p}^t)
\end{equation}

\vspace{1em}
\noindent \textbf{Inter-vehicle and Environment Repulsion:} 
In addition to the attractive force, vehicles are influenced by repulsive forces arising from neighboring vehicles and lane constraints, ensuring collision-free movements. The total repulsive force \(F_{\text{rep}}\) acting on the ego vehicle is defined by the negative gradient of a cumulative repulsive potential field \(U_{\text{total}}\):

\begin{equation}
F_{\text{rep}} = -\nabla U_{\text{total}}
\end{equation}

Where the total repulsive potential is decomposed into vehicle-based and lane-based components:

\begin{equation}
U_{\text{total}} = \sum_{j \in \Omega_n^t}(r_{\text{col}} k_{nj} e^{-\frac{\|\mathbf{r}_{nj}\|}{r_{\text{col}}}}) + \sum_{l \in \Lambda_n^t}U_{\text{line}, l}
\end{equation}

Specifically, for each neighboring vehicle \(j\), distinct repulsion parameters are employed depending on the vehicle's relative position: \(k_{np}\) for preceding vehicles, \(k_{nf}\) for following vehicles, and \(k_{nl}\) for lateral (side-by-side) vehicles. This distinction allows nuanced modeling of realistic inter-vehicle dynamics. Further, \( r_{\text{col}} \) denotes a scaling factor for the repulsive potential, given a target vehicle \( n \), the relative position of a neighboring vehicle  \( j \in \Omega_{n}^t \) is denoted as \( \mathbf{r}_{nj} = \mathbf{p}_n^t - \mathbf{p}_j^t \), \( \| \mathbf{r}_{nj} \| \) is the distance between these vehicles.

Lane-marking potentials, differentiating between crossable center lines and non-crossable boundary lines, are defined as:

\begin{equation}
U_{\text{line}, l} =
\begin{cases}
k_{cline} e^{-(d_{cline})^2}, & \text{for center lines} \\
k_{boundary}\frac{0.5}{(d_{boundary})^2}, & \text{for boundary lines}
\end{cases}
\end{equation}

Here, \( k_{cline}\) and \( k_{boundary}\) are adjustable strengths for center lines and boundary lines, respectively. In our social force model, center lane markings typically permit vehicle crossings, enabling maneuvers like overtaking or lane transitions. Consequently, we adopt an exponential decay function to represent their potential field, which exerts a moderate repulsive influence when vehicles closely approach these markings but rapidly weakens with increasing distance. On the other hand, boundary lines—such as road edges, medians, or barriers—serve as strict physical or regulatory constraints that vehicles must not cross. Therefore, we employ an inverse square potential for these non-crossable boundaries, ensuring a progressively stronger and persistent repulsive force as vehicles approach, which guarantees vehicles rigorously avoid collisions with these critical boundaries.

\subsection{Dual-Phase Physics-informed LLM for Safe Trajectory Planning}

\subsubsection{Memory Collection Phase}
The memory collection phase generates a knowledge-driven memory database that contains various driving scenarios and their corresponding social force model parameters. 
Specifically, it extracts key driving scenario data with the physics-informed prompts. These prompts explicitly embed social force modeling and physics-based constraints, guiding the LLM engine toward a comprehensive reasoning of the traffic scenarios. With their exceptional cognitive proficiency, LLMs can effectively interpret nuanced traffic situations, discern unsafe driving patterns, and apply learned physical principles to generate informed reasoning about vehicle dynamics. Moreover, explicit physical constraints from the social force model introduce robust physical realism into the LLM’s inferential process. This integration effectively transforms the ambiguous and hallucination-prone LLM-based trajectory prediction into a structured, parameter-driven optimization task enhanced by integrated verification mechanisms. The LLM-generated parameters are used by the social force model, along with corresponding goal states and predictions of neighboring vehicle behavior, to produce planned trajectories. These trajectories are validated by the reflection module, which examines and optimizes them based on comprehensive safety criteria. Validated trajectories are archived into the memory database, establishing a robust, reliable, and high-quality knowledge base to support the fast inference phase.

\begin{figure*}[bhtp]
  \begin{center}
  \includegraphics[width=.7\textwidth]{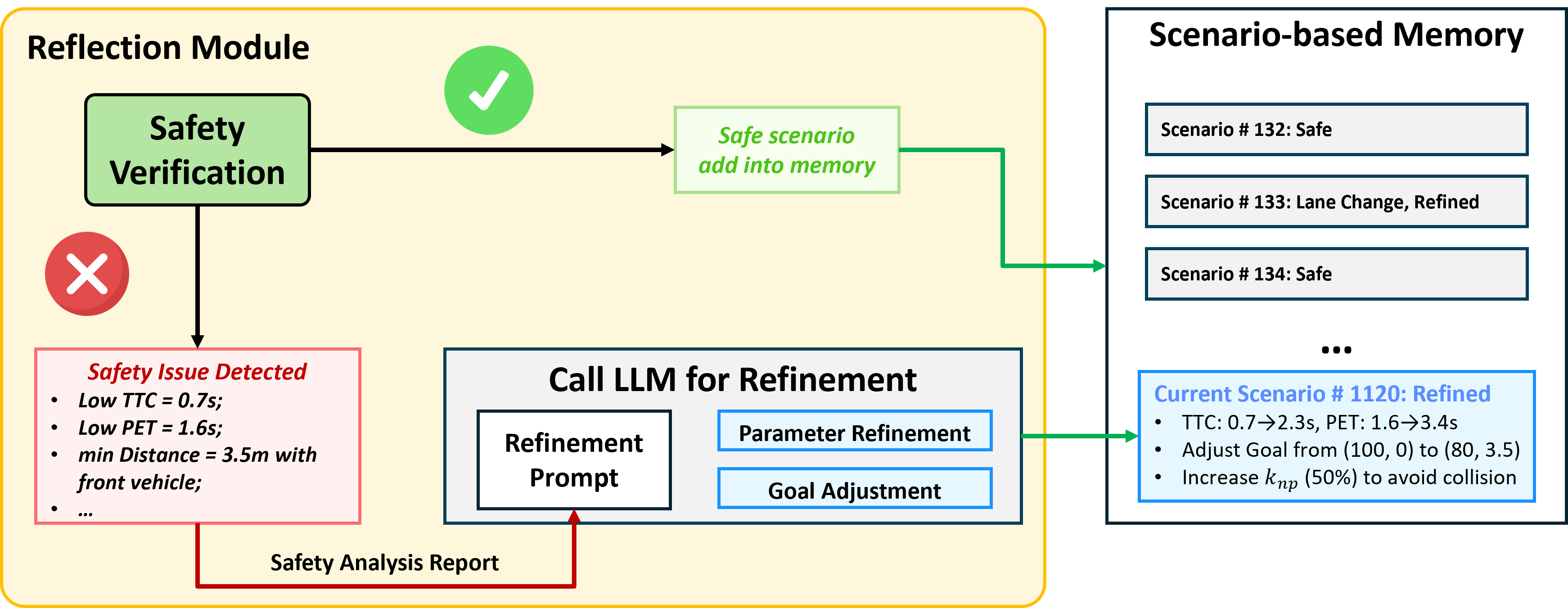}\\
  \caption{Reflection Module workflow. }\label{fig3_reflection}
  \end{center}
\end{figure*}

\textbf{Reflection Module.} This module evaluates and refines the trajectories as Fig.~\ref{fig3_reflection} shows. It identifies unsafe trajectories and refines potential conflicts to guarantee safe trajectory planning. Various safety metrics, including TTC, minimum neighbor-vehicle distance, and PET, are used in this module. When a trajectory violates predefined safety thresholds, such as a TTC shorter than 1.5 seconds or proximity less than 2.0 meters, this module generates a detailed safety analysis report. This report explicitly describes the precise circumstances of the safety concern, specifying factors such as time step, lane identification, and the exact neighbor vehicle interactions involved. This safety report subsequently informs an enhanced prompt, triggering another iteration of parameter refinement by the LLM to improve trajectory safety. Upon satisfying all safety requirements, the refined trajectories are saved into the memory database with scenario attributes, optimized parameters, and safety performance indicators. 

\textbf{Goal Adjustment.} 
The goal adjustment mechanism represents a key innovation within our reflective safety framework, enabling both proactive risk mitigation and enhanced trajectory planning. In the reflection module, goal adjustments are triggered when it detects unsafe scenarios characterized by small values of TTC or insufficient distance to neighboring vehicles. The goals are refined using two independent factors: a longitudinal adjustment factor and a lateral (lane) adjustment factor. Specifically, the longitudinal factor enables adaptive repositioning of the goal from aggressively forward-extended (value 0.0, permitting up to a 20\% increase in forward range) to conservatively shortened (value 1.0, decreasing the goal distance by up to 30\%). Differently, the lane factor spans from -1.0 (a complete shift to the adjacent left lane) to 1.0 (a complete shift to the adjacent right lane), with intermediate values indicating partial lateral adjustments. All adjusted goals maintain physical consistency by referencing and constraining adjustments relative to original positions. In the memory collection phase, when prompted by the detailed safety analysis report, the LLM generates refined goal positions and associated adjustment parameters based explicitly on identified collision categories (front, rear, or side impacts) and safety metrics. Successful adjustments are subsequently recorded in the memory bank, augmenting the knowledge database. During the fast inference phase, adjustment factors are swiftly determined via memory retrieval, eliminating the need for full reflective recalculations. In general practice, the system places greater emphasis on longitudinal adjustment to resolve most detected hazards, reserving lateral adjustments primarily for critical conditions where longitudinal adjustments alone provide insufficient safety margins. 

\subsubsection{Fast Inference Phase}
The Fast Inference phase is the deployment stage within the dual-phase LetsPi framework. This phase employs a specialized Fast Prompt strategy, exploiting the knowledge database built by the memory collection phase.  Unlike the comprehensive physical reasoning and reflective safety analyses performed during memory construction, the fast mode emphasizes rapid and actionable parameter inference tailored explicitly for practical, real-time driving scenarios.

The workflow begins by extracting critical features from new driving scenarios to retrieve analogous scenarios from the memory database. Leveraging these matched scenarios, the system constructs a concise prompt containing only essential contextual scenario information alongside historically optimal parameter guidance. The LLM processes this streamlined prompt and produces the necessary parameters without engaging in extensive reasoning. These parameters flow into the social force module, enabling consistent physics-based planning of vehicle trajectories while capitalizing on previously optimized and safety-validated parameter sets. 

\textbf{Fast Prompt.} The Fast Prompt substantially differs from the Physics-informed Prompt in both content and complexity, as shown in Fig.~\ref{fig2_prompt} b). The Fast Prompt is intentionally minimalistic. Specifically, it adopts a streamlined approach, selectively including only critical scenario characteristics, parameter recommendations derived from analogous, high-quality scenarios, and succinct, targeted guidance (e.g., "Approaching preceding vehicle—recommend elevated $k_{np}$ and adjust a conservative goal"). Moreover, the Fast Prompt omits the detailed reflection processes and intricate physical justifications, instead directly soliciting streamlined parameter outputs through a structured, easily parseable format optimized for rapid extraction and efficient inference.

\textbf{Scenarios Matching for Fast Prompting.} The efficiency of the fast inference phase depends on accurately identifying scenarios from the memory database. We define similarity between scenarios using a weighted feature-based approach that prioritizes safety-critical characteristics. Given a current scenario $S_{curr}$ with feature vector $\mathbf{f}_{curr}$ and a memory scenario $S_{mem}$ with feature vector $\mathbf{f}_{mem}$, we calculate their similarity as:

\begin{equation}
\text{Sim}(S_{curr}, S_{mem}) = \frac{1}{1 + D(S_{curr}, S_{mem})}
\end{equation}

where $D(S_{curr}, S_{mem})$ represents the weighted normalized feature distance:

\begin{equation}
D(S_{curr}, S_{mem}) = \sum_{k=1}^{n} w_k \cdot d_k(f_k^{curr}, f_k^{mem})
\end{equation}

The feature-specific distance function $d_k$ calculates the normalized difference between corresponding features:

\begin{equation}
d_k(f_k^{curr}, f_k^{mem}) = 1 - \min\left(1.0, \frac{|f_k^{curr} - f_k^{mem}|}{\max(f_k^{curr}, \epsilon)}\right)
\end{equation}

where $\epsilon = 10^{-5}$ prevents division by zero. We extract the following scenario features with their corresponding weights: number of vehicles in the scenario ($w = 1.0$), ego vehicle's initial speed in m/s ($w = 2.0$), boolean indicator for lane change maneuver ($w = 3.0$), minimum distance to any neighboring vehicle ($w = 2.5$), average distance to all neighboring vehicles ($w = 1.5$), and average speed of neighboring vehicles ($w = 1.0$). The weights are assigned based on feature importance to safety, with lane change scenarios and close-proximity situations receiving higher weights due to their safety-critical nature. After computing similarity scores for all scenarios in memory, we sort them in descending order and return the top-$k$ most similar scenarios:

\begin{equation}
\text{TopK}(S_{curr}) = \underset{S_i \in \mathcal{M}}{\text{argmax}_k} \{\text{Sim}(S_{curr}, S_i)\}
\end{equation}

where $\mathcal{M}$ represents the set of all scenarios in memory and $\text{argmax}_k$ returns the $k$ scenarios with highest similarity scores.

The fast inference phase significantly enhances efficiency without compromising on safety during deployment. By shortening the prompt length by approximately $50\%$ and removing explicit demands for comprehensive reasoning, this design reduces the token counts required for both input prompts and output responses. Additionally, this structural simplification enables the practical use of lighter-weight LLM models with fewer parameters, further amplifying computational speed and responsiveness. 

\section{Experiment}
\subsection{Detailed Setup}
We evaluate the proposed LetsPi using open-source LLMs as its foundational reasoning engines. Specifically, for the memory collection phase, we employ the Llama 3.1-8B model for sophisticated physics-informed reasoning and detailed driving scenario interpretation. The light-weight DeepSeek-R1 1.5B model is utilized for the fast inference phase. Experimental evaluations were performed on a Linux (Ubuntu) computing platform equipped with 64 GB of DDR5 RAM and powered by an NVIDIA GeForce RTX 4080 GPU with 16 GB of GPU memory. During inference, we maintained the LLM temperature parameter at 0 to produce deterministic outputs, while all other hyperparameters remained at their default settings to preserve reproducibility. Trajectories generated by our LetsPi model output at a resolution of 25 Hz, enabled by the integrated social force module, effectively translate complex LLM reasoning into practical, real-world trajectories.

\subsection{Data Description}
HighD dataset~\cite{krajewski2018highd} is used in this study for evaluation. HighD is a comprehensive, real-world dataset capturing naturalistic driving behaviors on highways. It comprises high-resolution drone-captured videos recorded at 25 Hz from 2017 to 2018 along a 420-meter segment of bidirectional highway near Cologne, Germany. It includes trajectories of approximately 110,000 vehicles, covering both passenger vehicles and heavy trucks, collectively traveling roughly 45,000 kilometers. The dataset encompasses six unique highway locations featuring diverse lane counts, road configurations, and on-ramp merging scenarios, thereby offering rich contextual variability critical for evaluating trajectory planning models.

For evaluation, trajectories are segmented into intervals of 8 seconds, where the first 3 seconds (equivalent to 75 frames) provide historical trajectory data and relevant state information. The LetsPi model generates trajectories for the next 5 seconds (125 frames). We implement a sliding-window segmentation strategy with a stride length of 20 frames. Additionally, any vehicle recorded within a 100-meter proximity of the ego vehicle throughout each scenario is treated as a neighbor vehicle (up to 8 neighbors), providing critical interaction context. From the fully processed dataset, we randomly select 10,000 representative scenarios, ensuring robust and statistically significant experimental evaluations.

\subsection{Evaluation on Real-world Driving Dataset}

\subsubsection{Baseline Models and Evaluation Metrics}
In our experimental framework, we comprehensively evaluated the proposed framework against three established baseline approaches across highway driving scenarios from the HighD dataset. The baseline models include:

\begin{itemize}
    \item \textbf{Social Force (SF) Model:} This baseline implements the classical social force paradigm where vehicle movements are governed by attractive forces toward goals and repulsive forces from lane marking and other vehicles. The model uses fixed parameters determined through offline optimization but lacks any adaptive capabilities. We employ the same social force architecture described in Section 3.4.
    
    \item \textbf{IDM:} This well-established car-following model regulates longitudinal vehicle control through sophisticated acceleration profiles. IDM incorporates safety distance considerations with parameters including desired velocity, minimum gap distance, comfortable acceleration, and deceleration rates. The model dynamically adjusts vehicle acceleration to maintain safe following distances that scale with velocity, providing realistic gap control and smooth deceleration when approaching slower vehicles.

    \item \textbf{Graph Neural Physics (GNP) \cite{gan2024goal}} Our previously developed Goal-based Neural Physics model represents a hybrid architecture combining physics-informed modeling with deep learning. GNP employs a two-stage prediction framework that first forecasts vehicle goals using multi-head attention mechanisms, then progressively constructs trajectories through neural-physical simulation. In contrast to our LLM-driven methodology, GNP explicitly utilizes driving data to train neural networks specifically designed to generate social force parameters. Furthermore, GNP lacks a built-in mechanism to perform adaptive refinement of potential goal positions based on safety-critical reflections, a critical advantage integrated in this research.
\end{itemize}

We evaluate model performance across 5 safety and 1 operational dimensions as follows.

\begin{itemize}
    \item \textbf{Safety - Average Minimum Time to Collision (TTC):} Temporal proximity to potential collisions between ego and its preceding vehicle, with values below 2s considered critically unsafe. Lower values indicate higher collision risk.
    
    \item \textbf{Safety - Collision Rate (CR):} Percentage of scenarios resulting in vehicle contact (distance $<$ 2.0m), directly quantifying catastrophic safety failures. 

    \item \textbf{Safety - Post Encroachment Time (PET):} Time interval between successive occupations of the same spatial point by different vehicles, capturing near-miss incidents through temporal separation. Lower values indicate higher collision risk.

    \item \textbf{Safety - Average Minimum Distance (minD):} The closest physical proximity between the ego vehicle and any other neighbor vehicles throughout the trajectory. Values below 2.0 meters are considered unsafe in highway scenarios. We report the average of the minimum distances observed across all test scenarios.

    \item \textbf{Safety - Low TTC Rate (LT):} Percentage of scenarios where minimum TTC falls below 2s, quantifying frequency of near-miss events requiring intervention.

    \item \textbf{Operational - Success Rate (SR):} Percentage of scenarios where the LLM-based model generates physically plausible trajectories reaching the destination within error bounds while maintaining safety constraints.
\end{itemize}

\subsubsection{Evaluation Results}
In Table~\ref{tab:comparison}, IDM shows the poorest performance, mostly due to its exclusive focus on longitudinal vehicle dynamics and insufficient capabilities for managing lane-change maneuvers in complex environments. The social force model is slightly better than IDM but is constrained by inflexible parameter settings, resulting in significant safety limitations, with a collision rate of 15.1\% and critically low TTC values identified in 10.1\% of scenarios. These outcomes reflect the model’s limited adaptability to dynamic traffic contexts but considerable safety concerns with repulsive forces. Conversely, the GNP model, which leverages physics-informed neural networks, markedly outperforms both IDM and the social force model, achieving a notably lower collision rate of 5.9\% and improved vehicle-spacing metrics. 

Significantly, our LetsPi framework consistently outperforms other baselines, achieving superior safety metrics in both phases. Specifically, the memory collection phase (denoted as LetsPi\_M) has the lowest collision rate at 2.8\% with large average minimum TTC values of 5.90 seconds. However, it has greater computational complexity and processing latency due to the emphasis on reasoning.
The fast inference phase (LetsPi\_F) delivers robust safety metrics alongside enhanced computational efficiency, achieving an impressive success rate of 99.6\%. This balance underscores that even with a simplified parameter inference approach, the knowledge-driven memory database effectively equips the LLM to generalize robustly and efficiently handle diverse driving situations.

\begin{table*}[!t]
    \caption{Comparison of Models on Key Safety Metrics}
    \label{tab:comparison}
    \centering
    \begin{tabular}{lccccccc}
        \toprule[1.2pt]  
        \textbf{Model} & 
        \textbf{\(\uparrow\) TTC (s)} & 
        \textbf{\(\downarrow\) CR (\%)} & 
        \textbf{\(\uparrow\) PET (s)} & 
        \textbf{\(\uparrow\) minD (m)} & 
        \textbf{\(\downarrow\) LT (\%)} & 
        \textbf{\(\uparrow\) SR (\%)} &
        \textbf{\(\downarrow\) Inference Time (s)} \\
        \midrule
        \textbf{SF}         & 4,13 & 8.9 & 1.31 & 17.12 & 11.5 & 100 & 0.1 \\
        \textbf{IDM}        & 3.94 & 15.1 & 0.92 & 13.49 & 10.1 & 100 & 0.1\\
        \textbf{GNP}        & 4.01 & 8.7 & 1.50 & 16.82 & 10.0 & 100 & 0.1\\
        \midrule 
        \textbf{LetsPi\_M}    & 5.90 & 2.8 & 2.80 & 16.75 & 6.0  & 97.0 & 6.18\\
        \textbf{LetsPi\_D}     & 5.98 & 3.4 & 2.65 & 17.92 & 8.0  & 99.6 & 3.12\\
        \bottomrule[1.2pt]  
    \end{tabular}
\end{table*}

\subsection{Visualization and Interpretability Analysis}
Two driving scenarios are used as examples to provide clear insights into the trajectory planning and refinement processes within the proposed LetsPi framework. Various visualization techniques are used to analyze vehicle interactions, LLM reasoning, and reflection results..

\begin{figure*}[bhtp]
  \begin{center}
  \includegraphics[width=.9\textwidth]{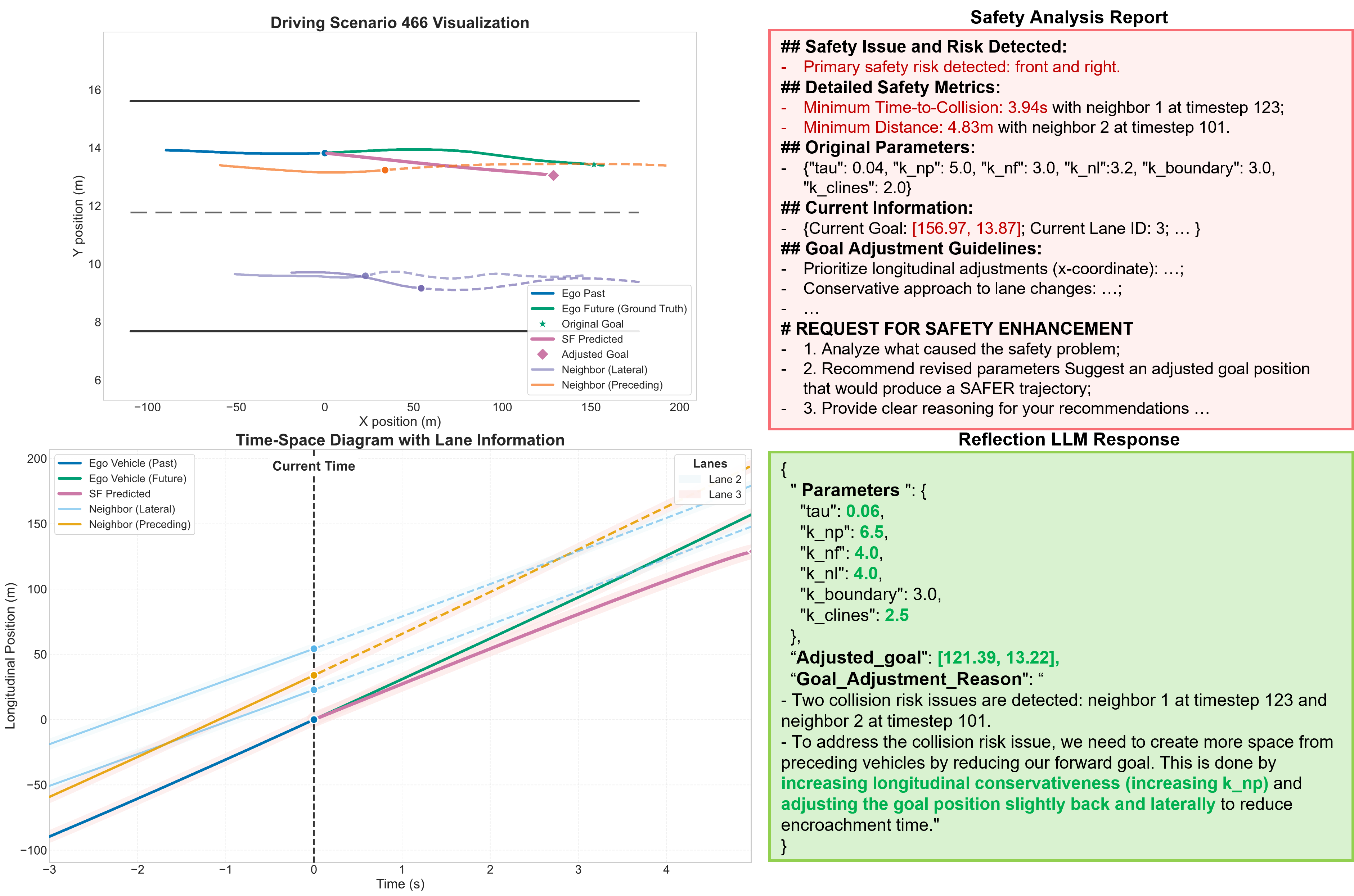}\\
  \caption{The visualization of a two-lane driving scenario. Elements include a trajectory diagram, a time-space diagram, a safety analysis report, and the response from the LLM. The trajectory diagram visualizes the past, predicted, and ground-truth trajectories of the ego-vehicle and neighboring vehicles. The time-space visualizes the relative position between the ego-vehicle and neighboring vehicles. The safety analysis report and reflection response provide detailed information on how the LLM refines its decisions based on safety criteria.}\label{fig4a}
  \end{center}
\end{figure*}

\begin{figure*}[bhtp]
  \begin{center}
  \includegraphics[width=.9\textwidth]{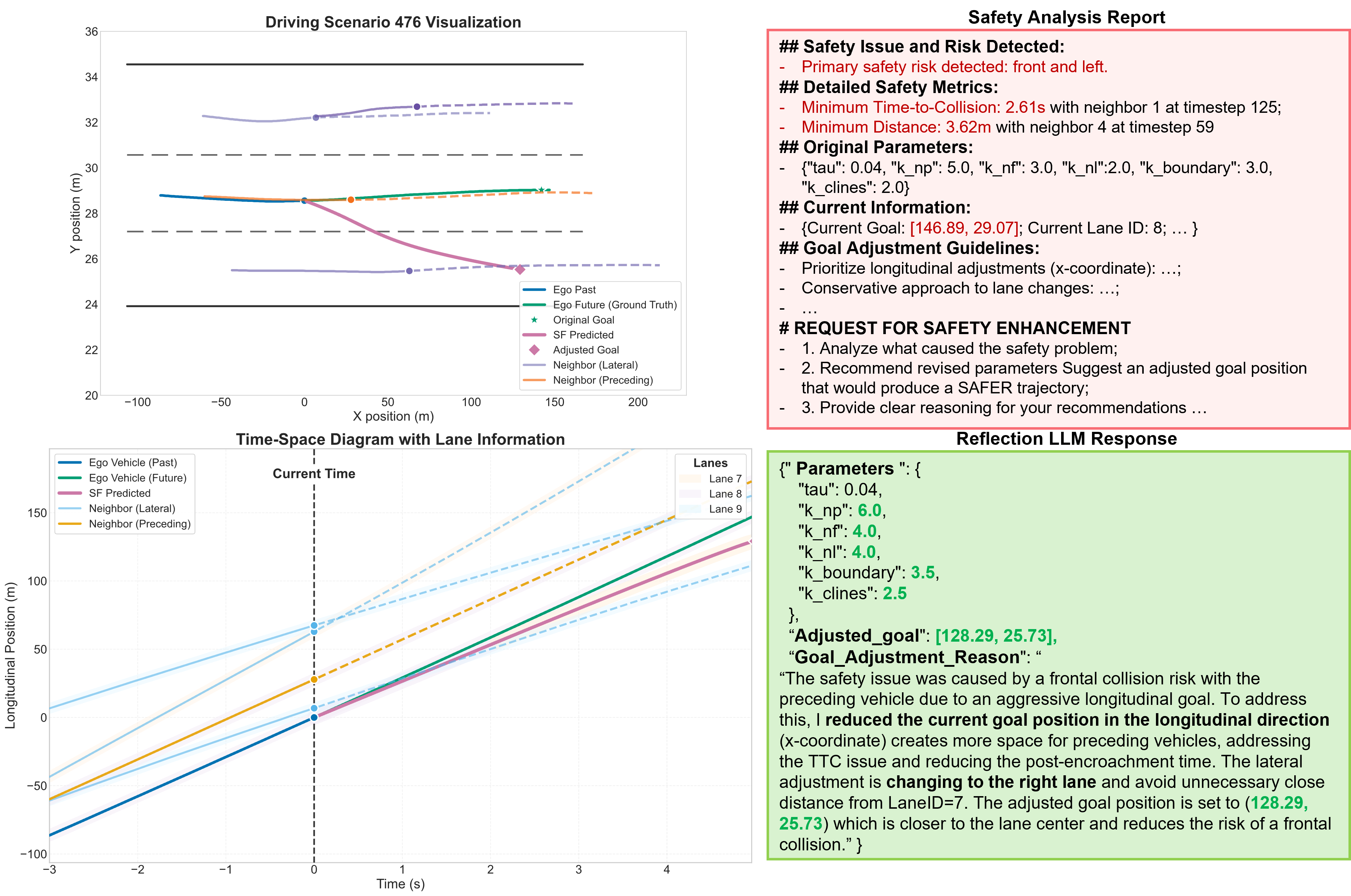}\\
  \caption{The visualization of a three-lane driving scenario. Elements include a trajectory diagram, a time-space diagram, a safety analysis report, and the response from the LLM. The trajectory diagram visualizes the past, predicted, and ground-truth trajectories of the ego-vehicle and neighboring vehicles. The time-space visualizes the relative position between the ego-vehicle and neighboring vehicles. The safety analysis report and reflection response provide detailed information on how the LLM refines its decisions based on safety criteria.}\label{fig4b}
  \end{center}
\end{figure*}

Fig.~\ref{fig4a} presents a two-lane driving context where the ego-vehicle follows another vehicle approximately 25 meters ahead, while simultaneously accompanied by two adjacent vehicles in the right lane. The time-space diagram shows an ongoing trend of decreasing inter-vehicle distance, which the reflection module identifies as a growing risk using TTC. The LLM proactively adjusts the trajectory goal longitudinally in a conservative manner, maintaining the current lane due to proximity to adjacent lane vehicles and the boundary line. Instead, it enhances lateral safety margins by adjusting lateral social force parameters.

Fig.~\ref{fig4b} represents a more challenging scenario, multiple neighboring vehicles closely positioned both longitudinally and laterally. The ego-vehicle maintains an initially small (20 meters) and diminishing gap with the preceding vehicle, while closely aligned laterally with another vehicle on the left. The reflection module identifies both of these relationships as critical factors. Therefore, the LLM refines the goal into the right lane, securing greater forward spacing and resolving lateral safety concerns. Interestingly, despite explicit guidance emphasizing longitudinal adjustments and restricting lane changes to severe scenarios, including “Prioritize longitudinal adjustments (x-coordinate)” or “Lane changes should be a last resort for severe safety concerns only”, the LLM still poses high lane-change maneuvers than human driving in HighD as an immediate and robust method for risk mitigation. In future research, alternative methods, such as reinforcement learning, should be explored to refine the driving style of LLM-based frameworks, potentially imitating and outperforming human-level performance.

\subsection{Ablation Study}

\textbf{Components.}
We performed comprehensive ablation experiments to examine the contributions of key modules using several variants as shown in Table~\ref{tab:ablation}. Specifically, "Base" denotes the complete dual-phase LetsPi model. The variant labeled "w/o Ref" indicates the removal of the reflection module. The "w/o GA" variant eliminates goal adjustment from the reflection module, retaining only the adjustments to the primary social force parameters. Finally, the "FS" variant represents a simplified model, removing both reflection and memory accumulation. It adopts a basic one-directional input-output model using a few-shot learning approach with three static exemplary scenarios embedded as prompts. 

Table~\ref{tab:ablation} highlights the critical role played by the reflection mechanism and the memory bank. Comparing "Base" and "w/o Ref," we find that without the reflection module, the model tends to store primarily general scenarios meeting safety criteria. As a result, the accumulated experiences inadequately prepare the model for diverse and challenging driving situations. A deeper comparison between "w/o Ref" and "w/o GA" illustrates that goal adjustments have a considerably stronger influence on trajectory optimization than mere modifications to social force parameters. Empirical observations confirm that the LLM utilizes goal adjustments to proactively mitigate potential collisions, either by adjusting distances to nearby vehicles or by opting for safer lane changes. Conversely, social force parameter adjustments typically influence detailed trajectory characteristics rather than fundamental trajectory goals. Thus, goal adjustments have a more immediate positive effect on critical safety metrics, reinforcing their significance within the reflection mechanism.

\begin{table}[!t]
    \caption{Ablation Study}
    \label{tab:ablation}
    \centering
    \begin{tabular}{lcccccc}
        \toprule[1.2pt]
        \textbf{Variant} & 
        \textbf{\(\uparrow\) TTC} & 
        \textbf{\(\downarrow\) CR} & 
        \textbf{\(\uparrow\) PET} & 
        \textbf{\(\uparrow\) minD} & 
        \textbf{\(\downarrow\) LT} & 
        \textbf{\(\uparrow\) SR} \\
        \midrule
        \textbf{w/o Ref}   & 4.33 & 8.0 & 1.84 & 16.77 & 11.0 & 99.7 \\
        \textbf{w/o GA}  & 4.18 & 4.7 & 1.86 & 17.01 & 10.7 & 99.7 \\
        \textbf{FS}  & 3.71 & 9.7 & 1.59 & 17.25 & 11.1 & 99.8 \\
        \textbf{Base}    & 5.98 & 3.4 & 2.65 & 17.92 & 8.0  & 99.6 \\
        \bottomrule[1.2pt]
    \end{tabular}
\end{table}

\textbf{Memory Size.}
Table~\ref{tab:mem_ablation} demonstrates a clear relationship between memory bank size and driving performance across all evaluated safety and operational metrics. As the memory capacity expands from 0\% (zero-shot) to 80\% (8,000 data points), we observe steady and consistent improvements in key safety indicators. These results validate that a continuously enriched memory database substantially improves model deployment performance. Moreover, the results indicate that the framework effectively utilizes prior experiences to refine its decision-making capabilities, with the performance gains directly proportional to the accumulated memory size.

\begin{table}[!t]
    \caption{Memory Ablation Study (Total = 10,000 Data Points)}
    \label{tab:mem_ablation}
    \centering
    \begin{tabular}{lcccccc}
        \toprule[1.2pt]
        \textbf{Memory} & 
        \textbf{\(\uparrow\) TTC} & 
        \textbf{\(\downarrow\) CR} & 
        \textbf{\(\uparrow\) PET} & 
        \textbf{\(\uparrow\) minD} & 
        \textbf{\(\downarrow\) LT} & 
        \textbf{\(\uparrow\) SR} \\
        \midrule
        0 (0\%)  & 3.71 & 9.7 & 1.59 & 17.25 & 11.1 & 99.8 \\
        100 (1\%) & 3.67 & 11.0 & 1.67 & 16.79 & 15.8 & 99.3 \\
        1000 (10\%) & 4.18 & 6.5 & 2.25 & 17.85 & 9.2 & 99.5 \\
        5000 (50\%) & 5.77 & 4.8 & 2.46 & 17.93 &  8.5 & 99.5 \\
        8000 (80\%) & 5.98 & 3.4 & 2.65 & 17.92 & 8.0  & 99.6 \\
        \bottomrule[1.2pt]
    \end{tabular}
\end{table}

\textbf{Different LLM Combinations.}
We evaluated various LLM combinations in the proposed framework. In general, the memory collection phase benefits from employing larger-scale models, enabling deeper comprehension of driving scenarios, physical models, and facilitating thorough accumulation of high-quality decision data through the refinement module. In contrast, the fast inference phase prioritizes smaller-scale LLM architectures, chosen explicitly for rapid analysis and inference efficiency, effectively referencing stored memory to plan trajectories. Table~\ref{tab:llm_combos} indicates the superior performance of larger-parameter models, such as DeepSeek-R1:14B, over the Llama 3.1:8B model in overall metrics for the memory collection phase. Likewise, in the fast inference phase, the Llama 3.2:3B model consistently outperforms smaller alternatives, such as DeepSeek-R1:1.5B. 


Moreover, the two-phase design considerably improves real-time practicality. For 5-second future trajectory generation, smaller-scale LLMs (Llama 3.2:3B, DeepSeek-R1:1.5B) achieve practical update rates of approximately 0.2–0.3 Hz, closely aligning with the realistic demands of real-world driving assistance systems. Future research incorporating even more powerful LLMs during the memory collection phase and advanced distillation techniques for the fast inference phase holds significant potential to further enhance both trajectory quality and computational efficiency, enabling robust deployment in practical driving-assistance scenarios.

\begin{table*}[!t]
    \caption{Combinations of LLM Models and Their Performance}
    \label{tab:llm_combos}
    \centering
    \begin{tabular}{l c l c | c c c c c c}
        \toprule[1.2pt]
        \textbf{Mem LLM} & \textbf{Mem Inf(s)} & 
        \textbf{Fast LLM} & \textbf{Fast Inf(s)} & 
        \textbf{\(\uparrow\) TTC (s)} & \textbf{\(\downarrow\) CR (\%)} & \textbf{\(\uparrow\) PET (s)} & \textbf{\(\uparrow\) minD (m)} & 
        \textbf{\(\downarrow\) LT (\%)} & \textbf{\(\uparrow\) SR (\%)} \\
        \midrule
        deepseek-r1:14b & 17.92 & llama3.1:8B & 5.35 & \textbf{6.13} & \textbf{2.4} & \textbf{3.20} & 16.98 & \textbf{5.7} & 93.5 \\
        deepseek-r1:14b & 18.36 & llama3.2:3B & \textbf{3.07 }& 6.01 & 3.0 & 3.15 & \textbf{18.07} & 6.0 & 93.7 \\
        deepseek-r1:14b & 18.05 & deepseek-r1:1.5b & 3.16 & 5.71 & 3.1 & 2.25 & 17.32 & 7.8 & 89.0 \\
        \midrule
        llama3.1:8B & 6.34 & llama3.1:8B & 5.38 & 5.97 & 2.8 & 2.79 & 17.34 & 7.4 & 99.1 \\
        llama3.1:8B & 6.18 & llama3.2:3B & 3.12 & 5.98 & 3.4 & 2.65 & 17.92 & 8.0  & \textbf{99.6} \\
        llama3.1:8B & 6.15 & deepseek-r1:1.5b & 3.36 & 5.61 & 3.2 & 2.45 & 17.80 & 8.2 & 95.0 \\
        \bottomrule[1.2pt]
    \end{tabular}
\end{table*}

\section{Conclusion}
In this paper, we propose LetsPi, a physics-informed, dual-phase, knowledge-driven framework for safe, human-like trajectory planning. This hybrid framework integrates the LLM's strong reasoning capability with the social force model's physical dynamics. Moreover, the dual-phase design balances in-depth reasoning and computing efficiency, enhancing the applications of the framework in real-world scenarios.

To address the challenges of hallucinations and uncertainty in the LLM's outputs, we present a physics-aware prompting structure, allowing the LLM to infer key trajectory parameters and destination goals, which the social force model subsequently translates into physically realistic driving trajectories. Additionally, LetsPi’s unique architecture includes a reflective Memory Collection phase, where trajectory predictions are iteratively refined and stored using various surrogate safety measures as validated driving knowledge. A separate Fast Inference phase is designed for efficient, reliable real-time trajectory planning. Extensive experiments on real-world highway driving data demonstrate LetsPi’s clear superiority over traditional model-based and learning-based approaches, consistently outperforming these baselines across various criteria. Our extensive ablation analyses confirm the essential role of the reflective goal-adjustment mechanism, enhancing trajectory safety beyond standard few-shot learning approaches. Additionally, LetsPi’s performance scales effectively with accumulated driving experiences, similar to human drivers. Analysis of various LLMs reveals a practically effective deployment strategy: larger-scale models for the memory accumulation phase and smaller-scale models for rapid inference. 

While LetsPi exhibits strong performance, it currently has limitations we aim to overcome in future research. First, due to the absence of visual modality in the HighD dataset, future work will integrate VLM and richer multimodal data. Second, the current dependence on precise goal and neighbor trajectory estimations of the social force model motivates exploration of hybrid models combining LLM or VLM with other physics or rule-based frameworks. Finally, we envision implementing LetsPi’s dual-phase structure within a vehicle-cloud paradigm, leveraging cloud-based fine-tuned models for memory collection and lightweight, vehicle-deployed models for rapid trajectory inference.

\bibliographystyle{IEEEtranN}
\bibliography{main.bib}

\newpage

\begin{IEEEbiography}[{\includegraphics[width=1in,height=1.25in,clip,keepaspectratio]{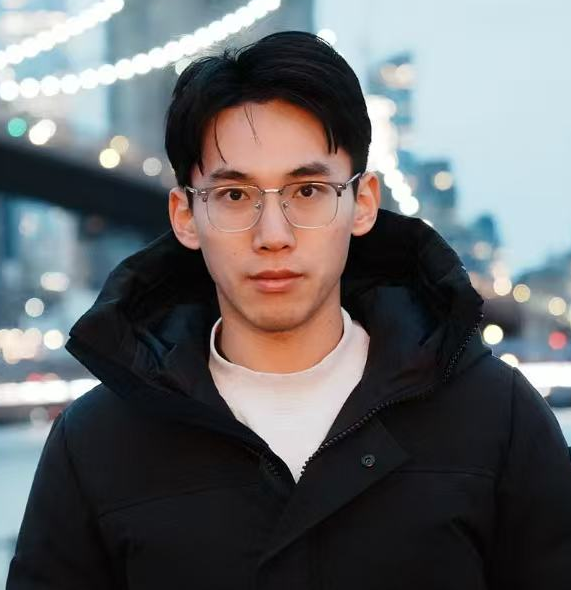}}]{Rui GAN}
received his B.S. and M.S. degrees in Traffic Engineering from Southeast University in 2020 and 2022, respectively. He is currently pursuing a Ph.D. in Civil and Environmental Engineering at the University of Wisconsin-Madison. His research focuses on AI in intelligent connected autonomous vehicles and infrastructures, Vehicle motion prediction and planning and LLM-empowered autonomous driving systems.
\end{IEEEbiography}

\begin{IEEEbiography}[{\includegraphics[width=1in,height=1.25in,clip,keepaspectratio]{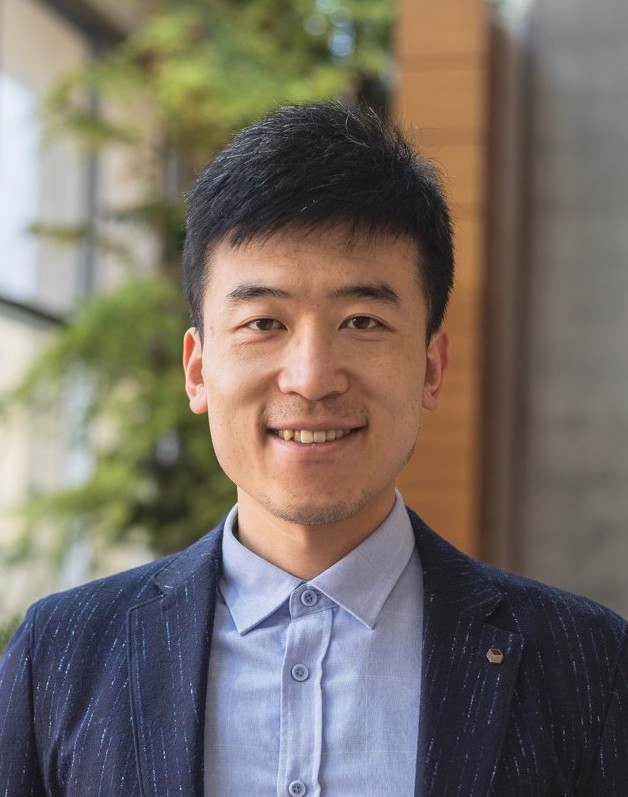}}]{Pei LI} is a Scientist in the Department of Civil and Environmental Engineering at the University of Wisconsin-Madison. He received his Ph.D. in Civil Engineering with a focus on Transportation Engineering from the University of Central Florida in 2021, after which he served as a Postdoctoral Research Fellow at the University of Michigan Transportation Research Institute. His research interests include transport safety, smart mobility, human factors, machine learning, connected and automated vehicles, and digital twins.
\end{IEEEbiography}

\begin{IEEEbiography}[{\includegraphics[width=1in,height=1.25in,clip,keepaspectratio]{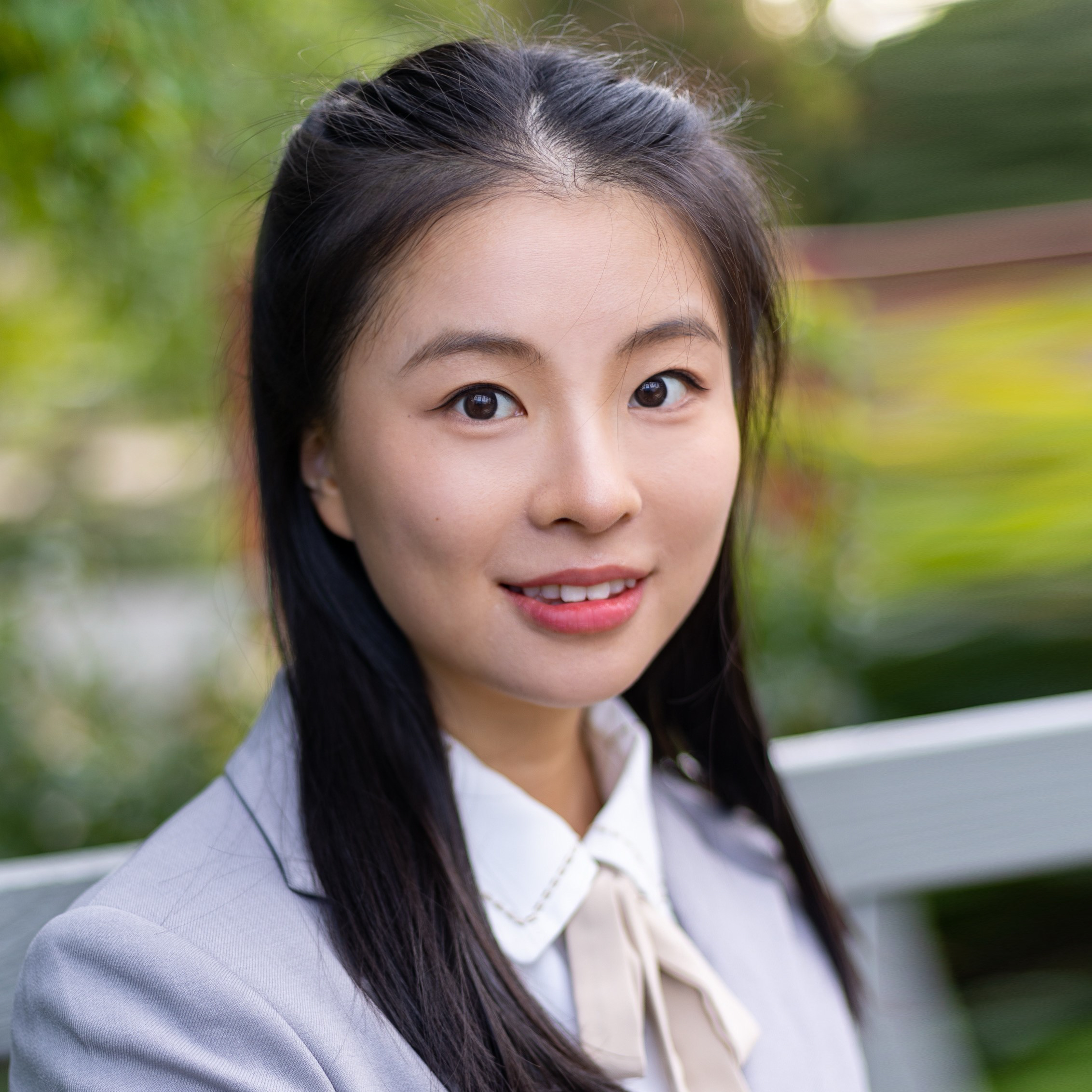}}]{Keke LONG}
is a Postdoc Research Associate at University of Wisconsin, Madison, USA. She obtained her Ph.D. degree in civil engineering in University of Wisconsin, Madison, USA in 2024, Master's degree in Traffic Engineering from Tongji University, China in 2021, and B.S. degree in Traffic Engineering from Changan University, China in 2018. Her main research interests are connected and autonomous vehicles and artificial intelligence in transportation. 
\end{IEEEbiography}

\begin{IEEEbiography}[{\includegraphics[width=1in,height=1.25in,clip,keepaspectratio]{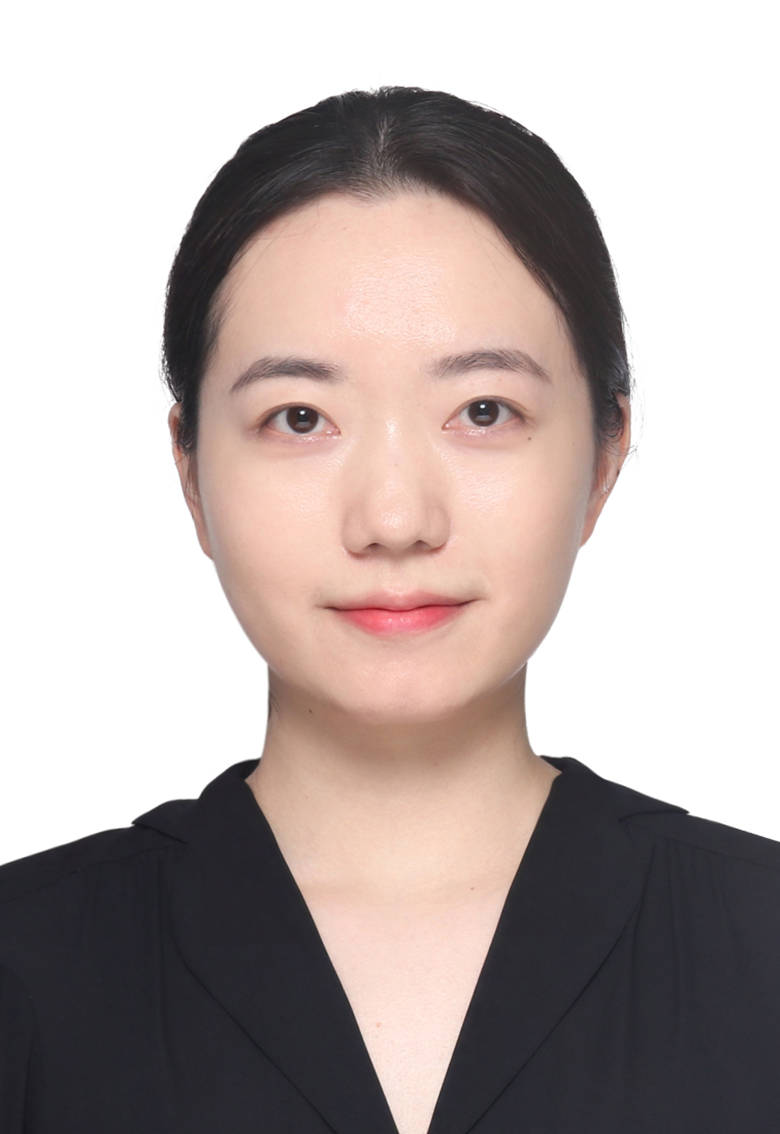}}]{Junyi MA}
is a Ph.D. student at University of Wisconsin, Madison, USA. She obtained her Master's degree in Construction Economics and Management from University College London, UK in 2020, and Bachelor's degree in Economics and Statistics from University College London, UK in 2019. Her main research interests are connected and autonomous vehicles and artificial intelligence technologies in the transportation research.
\end{IEEEbiography}

\begin{IEEEbiography}[{\includegraphics[width=1in,height=1.25in,clip,keepaspectratio]{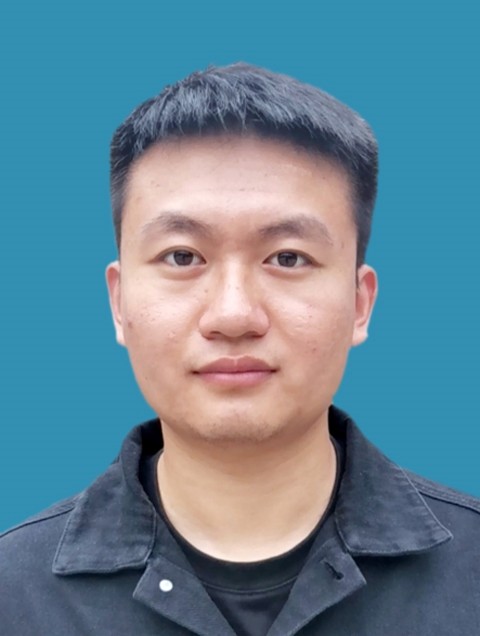}}]{Bocheng AN}
received the M.S. degree from Southeast University, Nanjing, China, in 2023, where he is currently pursuing the Ph.D. degree with the School of Transportation. His research primarily focuses on the application of artificial intelligence technologies in the transportation research, including vehicle trajectory prediction and trajectory planning.
\end{IEEEbiography}

\begin{IEEEbiography}[{\includegraphics[width=1in,height=1.25in,clip,keepaspectratio]{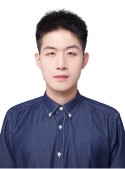}}]{Junwei YOU}
received the M.S. degree in Civil and Environmental Engineering from Northwestern University in 2022. He is currently a Ph.D. student in Civil and Environmental Engineering at University of Wisconsin-Madison. His research interests are autonomous driving, foundation models, generative AI, and intelligent transportation systems.
\end{IEEEbiography}

\begin{IEEEbiography}[{\includegraphics[width=1in,height=1.25in,clip,keepaspectratio]{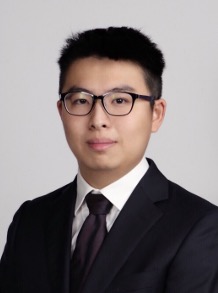}}]{Keshu WU}
is a postdoctoral research associate at Texas A\&M University. He receives his Ph.D. in Civil and Environmental Engineering from the University of Wisconsin-Madison in 2024. He also holds an M.S. degree in Civil and Environmental Engineering from Carnegie Mellon University in 2018 and an M.S. degree in Computer Sciences from the University of Wisconsin-Madison in 2022. He completed his B.S. in Civil Engineering at Southeast University in Nanjing, China in 2017. His research interests include the application and innovation of artificial intelligence and deep learning techniques in connected automated driving, intelligent transportation systems, and digital twin modeling and simulation.
\end{IEEEbiography}

\begin{IEEEbiography}[{\includegraphics[width=1in,height=1.25in,clip,keepaspectratio]{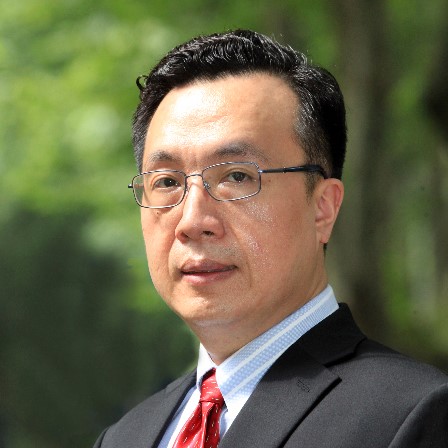}}]{Bin RAN}
is the Vilas Distinguished Achievement Professor and Director of ITS Program at the University of Wisconsin at Madison. Dr. Ran is an expert in dynamic transportation network models, traffic simulation and control, traffic information system, Internet of Mobility, Connected Automated Vehicle Highway (CAVH) System. He has led the development and deployment of various traffic information systems and the demonstration of CAVH systems. Dr. Ran is the author of two leading textbooks on dynamic traffic networks. He has co-authored more than 240 journal papers and more than 260 referenced papers at national and international conferences. He holds more than 20 patents of CAVH in the US and other countries. He is an associate editor of Journal of Intelligent Transportation Systems.
\end{IEEEbiography}

 




\vfill

\end{document}